\documentclass{article}
\usepackage{spconf,amsmath,graphicx}
\usepackage{amssymb, bbm, setspace, mathrsfs, color, listings, graphics, multicol, dcolumn, dsfont, subcaption, float, mathabx, booktabs}
\usepackage{caption}
\usepackage{tabularx}
\usepackage{algorithm}
\usepackage[noend]{algpseudocode}
\usepackage{multirow}
\usepackage{hyperref}
\usepackage{breakurl}
\usepackage{enumitem}

\DeclareMathOperator*{\argmin}{argmin}

\newcommand{\rpm}{\raisebox{.2ex}{$\scriptstyle\pm$ }}

\title{POLA: Online Time Series Prediction by Adaptive Learning Rates}
\name{Wenyu Zhang}
\address{Institute for Infocomm Research, Singapore}

\begin{document}
\ninept
\maketitle
\begin{abstract}

Online prediction for streaming time series data has practical use for many real-world applications where downstream decisions depend on accurate forecasts for the future. Deployment in dynamic environments requires models to adapt quickly to changing data distributions without overfitting. We propose POLA (\textbf{P}redicting \textbf{O}nline by \textbf{L}earning rate \textbf{A}daptation) to automatically regulate the learning rate of recurrent neural network models to adapt to changing time series patterns across time. POLA meta-learns the learning rate of the stochastic gradient descent (SGD) algorithm by assimilating the prequential or interleaved-test-then-train evaluation scheme for online prediction. We evaluate POLA on two real-world datasets across three commonly-used recurrent neural network models. POLA demonstrates overall comparable or better predictive performance over other online prediction methods.
\end{abstract}

\section{INTRODUCTION}

As data collection becomes more accessible, there is an increase in the amount of data available for analysis. In streaming settings, data is collected by continuously monitoring a system. A key challenge for model deployments in dynamic environments is concept drift, where the joint distribution of predictor and response variables change across time, such that static models can become obsolete~\cite{parisi2019review}. Online learning techniques continuously update the model as new data streams in to adapt to the current state of the system~\cite{mcmahan2011equivalence, xiao2009dual}, in some cases with the constraint that samples are immediately discarded after being learned due to reasons such as limited storage, privacy and to reduce interference from outdated samples~\cite{mcmahan2017surveyadaptive}.

In this work, we focus on online time series prediction which is studied and applied in a wide range of domains such as in traffic monitoring, climate research and finance~\cite{Miyaguchi2019CograCS, liu2020rbf, Lu2017TimeSP, anava2013online}. Predictions can be used for downstream tasks such as anomaly detection and to aid decision making~\cite{ahmad2017htm, chen1993outliers}. Unlike in the traditional online learning framework, temporal correlations means that observations cannot assumed to be independently and identically distributed (i.i.d.). In this work, we also focus on deep recurrent neural network (RNN) models which are popularly used to capture complex temporal correlations~\cite{krstanovic2017ensemble}. Given sufficient data at training, these deep models are shown to have competitive prediction performance over conventional statistical time series models without excessive pre-processing and feature engineering~\cite{hewalage2020rnn, Zhang2020MultilabelPI}. 

\begin{figure}
    \centering
    \includegraphics[width=\linewidth]{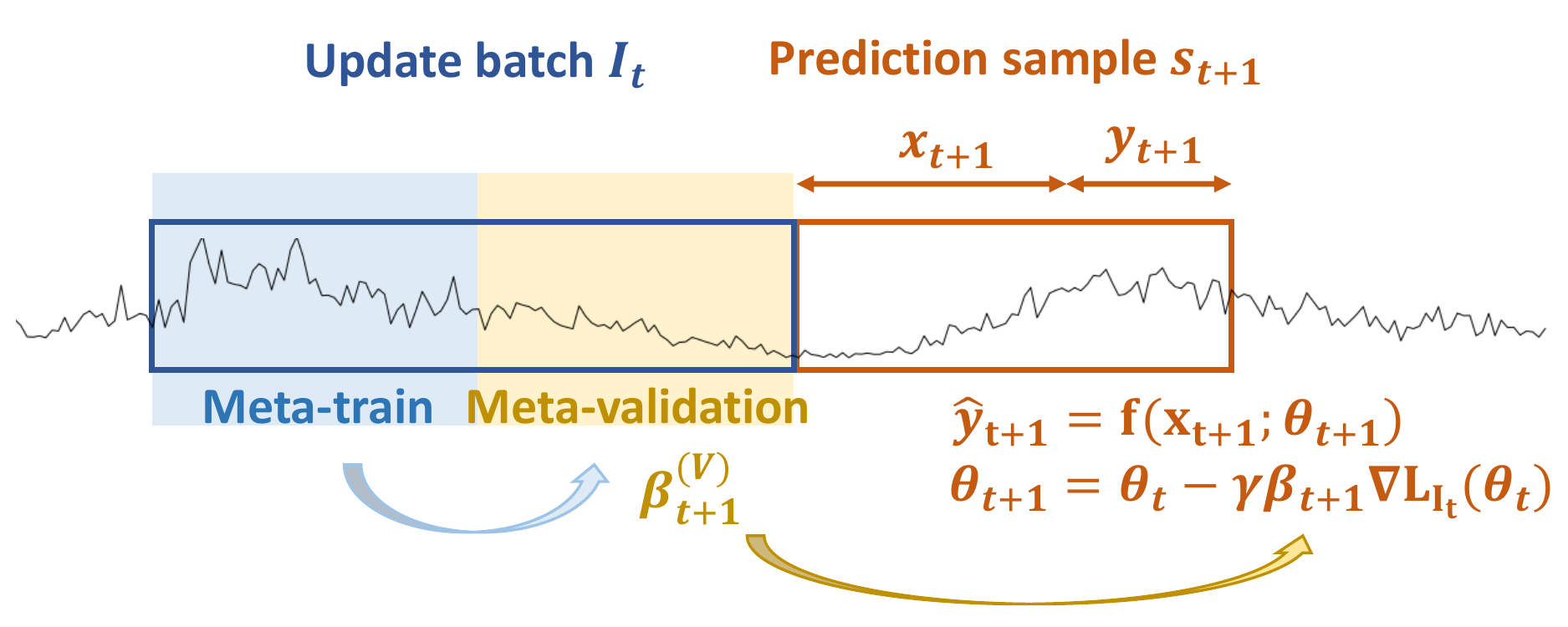}
    \caption{Overview: POLA adapts the learning rate of SGD update through factor $\beta_{t+1}\in [0,1]$ where the factor is meta-learned with the most recent data batch by assimilating online evaluation.}
    \label{fig:overview}
\end{figure}

Deep models are typically learned by gradient descent algorithms. Most works design algorithms for stationary environments with i.i.d. samples~\cite{mcmahan2017surveyadaptive}, and others that do consider time series prediction require Hessian computations~\cite{Miyaguchi2019CograCS, schaul2013sgdv} that are computationally intensive for neural networks with large number of parameters. We propose POLA (\textbf{P}redicting \textbf{O}nline by \textbf{L}earning rate \textbf{A}daptation) to 
\begin{itemize}[noitemsep]
    \item Adapt quickly in dynamic environments without overfitting to current system state or noisy samples, by
    \item Automatically scheduling the online learning rate of the stochastic gradient descent (SGD) algorithm by assimilating online evaluation.
\end{itemize}
Figure~\ref{fig:overview} provides an overview of the proposed method. POLA estimates the learning rate with a novel meta-learning~\cite{hospedales2020meta} procedure that assimilates the interleaved-test-then-train evaluation scheme for online prediction. Since SGD update only uses the most recently observed batch of data, POLA operates under the constraint that samples are discarded after being learned, which further makes it suitable for light-weight applications. We test our proposed approach on two publicly available real-world datasets across three commonly-used recurrent neural network architectures.

\section{RELATED WORKS}

Gradient descent algorithms are commonly used to train neural network models online by constantly tracking a specified performance measure such as prediction error, and are also applied to traditional statistical methods such as the autoregressive integrated moving average models~\cite{anava2013online, Liu2016OnlineAA}. The learning rate can be adapted to achieve better convergence, avoid overfitting to noisy samples or to automatically adjust to shifts in data distribution~\cite{Miyaguchi2019CograCS, schaul2013sgdv, bartlett2007adaptive, George2006AdaptiveSF, Jenni2018DeepBL}. 
RMSprop \cite{rmsprop} is a popular algorithm that scales the learning rate by moving average of squared gradients based on the intuition that the magnitude of each weight update should be similar regardless of the actual gradient magnitude. In POLA, instead of explicitly applying an inductive bias into learning rate scaling, we propose learning this scaling directly from the data itself. 
For adaptation in non-stationary environments, previous works apply simplifying assumptions on the loss function and Hessian computations~\cite{Miyaguchi2019CograCS, schaul2013sgdv}, however second-order derivative calculation can be expensive and the resulting learning rates are subject to instability if the approximated Hessian is ill-conditioned. Other works adapt the learning rate to reduce the adverse effect of outliers by monitoring distributional properties at neighboring data points~\cite{guo2016robust, yang2017robust}.

Popular alternatives to gradient descent algorithms include the online sequential extreme learning machine~\cite{huang2005oselm}, which updates by recursive least squares. Extension to time series modeling includes variants with RNN structure~\cite{park2017orelm}, and moderating the extent of model update through a confidence coefficient based on the difference between prediction error and a set threshold~\cite{Lu2017TimeSP}. Other works optimize for the model parameters by genetic programming~\cite{wagner2007genetic}, alternating parameter and hyperparameter optimization~\cite{zhan2018ridge}, or use concept drift detection to detect data distribution changes through specified rules and subsequently initiate model updates~\cite{liu2020rbf, aljundi2019continual, rao2019unsupervised}. However, incorporating concept drift detection as a stand-alone step can introduce additional errors, and fixed rules may be too rigid or restrictive for real-world applications. POLA builds on the SGD algorithm so that the model is updated directly based on prediction loss without intermediate concept drift detection steps.

\section{PROBLEM SETUP}

For a process $\{Z_t\}$, we denote the observation at time $t$ as $z_t \in \mathbb{R}^d$ with dimensionality $d$. At each time $t$, the prediction task is to use $x_t = \left[z_{t-m+1}, \dots, z_t\right]$, a historical input sequence of length $m$, to predict $y_t = \left[z_{t+1}, \dots, z_{t+n}\right]$, an output forecast sequence for the next $n$ time steps. We denote the prediction sample at time $t$ as $s_t = (x_t, y_t)$, that is, samples are obtained by a sliding window of stride 1 on the sequence $\{z_t\}$. Note that the samples observed up until time $t$ are $\{s_i\}_{i=m}^{t-n}$. The samples $\{s_i\}_{i=t-n+1}^t$ are not observed yet since they consist the value $z_{t+1}$. 

Let $f$ be a prediction model parameterized by $\theta_t$ at time $t$. Then the model predicts
\begin{equation}
    \hat{y}_t = f(x_t; \theta_t)
\end{equation}
and it incurs a loss $\ell_t(\theta_t; x_t, y_t)$ such as the squared prediction error $\|y_t - f(x_t; \theta_t)\|_2^2$. At the next time step $t+1$, the value $z_{t+1}$ and consequently the sample $s_{t-n+1}$ is observed and can be used to update model parameters to $\theta_{t+1}$. For completeness of notations, $\ell_t=0$ for $1\leq t < m$ when the number of observations is insufficient to form samples of the desired length.

Let $b$ be the number of most recent samples that can be stored and learned at any time $t$, we also refer to $b$ as the online batch size. We denote the samples available for model update at time $t$ as $U_t = \{s_i\}_{i\in I_t}$ where $|I_t|\leq b$. In this paper, we only consider the common setting where the model is updated with data chunks of a fixed size~\cite{guo2016robust, huang2005oselm, park2017orelm}, that is, when $|I_t|=b$ and explicitly $I_t = \{t-n-b+1, \dots, t-n\}$. We assume $b>1$ since batch update is generally less noisy than single sample update, and it also reduces the amount of computation required for frequent updates which may be inhibiting for light-weight applications.

\section{PROPOSED METHOD: POLA}

\subsection{Formulation}

The objective of online prediction is to learn model parameters or hypothesis to minimize the loss function $\ell_{t+1}$ at the future time step:
\begin{equation}
    \theta^*_{t+1} = \argmin_{\theta} \ell_{t+1}(\theta; x_{t+1}, y_{t+1})
\end{equation}
where $y_{t+1}$ is unobserved when we estimate for $\theta^*_{t+1}$ at time $t$.

Since we estimate $\theta^*_{t+1}$ based on SGD, assuming $|I_t|=b$ such that model parameters are updated at step $t$ with online batch size $b$, then the estimated parameters $\theta_{t+1}$ take the form
\begin{align}
    \theta_{t+1} &= \theta_{t} - \gamma \beta_{t+1} \nabla L_{I_t}(\theta_{t}) \label{eqn:sgd_adaptive} \\
    \text{where } L_{I_t}(\theta_t) &= \sum_{i \in I_t} \ell_{i}(\theta_{t}; x_i, y_i) \nonumber
\end{align}
where we introduce a maximum learning rate $\gamma$ and an additional factor $\beta_{t+1} \in [0,1]$ to represent the overall learning rate as $\gamma_{t+1} = \gamma \beta_{t+1}$. Such decomposition is also used in~\cite{guo2016robust}. When $|I_t|<b$ such that there are insufficient samples for a batch update,  then $\theta_{t+1} = \theta_{t}$. The intuition for $\beta_{t+1}$ is that the learning rate should be small if the update batch is not useful in helping the model adapt.

The adaptive update can be formulated as bilevel optimization for both $\theta_{t+1}$ and $\beta_{t+1}$:
\begin{align}
    \theta_{t+1}, \beta_{t+1} &= \argmin_{\theta, \beta} \ell_{t+1}(\tilde{\theta}; x_{t+1}, y_{t+1}) \label{eqn:bilevel}\\
    \text{subject to } \tilde{\theta} &= \argmin_{\theta} \left\lbrace \gamma\beta L_{I_t}(\theta) - (1/2) \|\theta-\theta_{t}\|^2_2 \right\rbrace \nonumber
\end{align}
and approximation of $L_{I_t}(\theta)$ with a first-order Taylor expansion about the point $\theta_{t}$ gives the adaptive SGD update in Equation~\ref{eqn:sgd_adaptive} for $\theta_{t+1}$, following proximal view of gradient descent~\cite{polson2015proximal}. We estimate $\beta_{t+1}$ through a meta-learning procedure in Section~\ref{sec:metalearn}.

\subsection{Meta-learning For Learning Rate}
\label{sec:metalearn}

The lower level objective in Equation~\ref{eqn:bilevel} corresponds to the training set and the upper level objective corresponds to the test set. We create a meta-training set $U_t^{(T)}$ and meta-validation set $U_t^{(V)}$ by splitting $U_t$, the update batch of samples indexed by $I_t$, as
\begin{align}
    U_t^{(T)} &= \{s_i\}_{i \in I_t^{(T)}} \text{ and }
    U_t^{(V)} = \{s_i\}_{i \in I_t^{(V)}} \label{eqn:split}\\
    \text{ where } & I_t^{(T)} = \{t-n-j\}^{b-1}_{j=\lceil b/2 \rceil} \text{ and } I_t^{(V)} = \{t-n-j\}^{\lceil b/2 \rceil - 1}_{j=0}\nonumber 
\end{align}
such that the meta-training and meta-validation sets are a proxy to the training and testing procedure. The bilevel optimization in the meta-stage is then
\begin{align}
    \theta^{(V)}_{t+1}, \beta^{(V)}_{t+1} &= \argmin_{\theta, \beta} L_{I^{(V)}_t}(\tilde{\theta}) \label{eqn:bilevel_meta} \\
    \text{subject to } \tilde{\theta} &= \argmin_{\theta} \left\lbrace \gamma\beta L_{I^{(T)}_t}(\theta) - (1/2) \|\theta-\theta_t\|^2_2 \right\rbrace \nonumber 
\end{align}
and we approximate $\beta_{t+1}$ with $\beta^{(V)}_{t+1}$. Empirically, we set $\beta_{t+1}$ as the moving average of the most recently computed $\beta^{(V)}$'s.

\subsection{Implementation}

We present two versions of POLA corresponding to two approaches to estimate $\beta_{t+1}^{(V)}$, namely POLA-FS and POLA-GD, in Algorithm~\ref{alg:pola}. POLA-FS searches for the optimal learning rate $\gamma_t = \gamma \beta_{t+1}^{(V)}$ in a finite set of candidates denoted $\mathcal{C}$. Replacing the solution to the lower level objective in Equation~\ref{eqn:bilevel_meta} with the SGD update, we get
\begin{equation}
    \beta^{(V)}_{t+1} = \argmin_{\beta \text{ s.t. } \gamma\beta \in \mathcal{C}} L_{I^{(V)}_t}\left(\theta_{t} - \gamma \beta \nabla L_{I_t^{(T)}}(\theta_{t})\right).
    \label{eqn:pola-fs}
\end{equation}
The version POLA-GD optimizes for $\beta_{t+1}^{(V)}$ by gradient descent while freezing all other model parameters. Two additional hyperparameters define the gradient descent procedure, namely the number of update steps $k$ and the learning rate $\eta$. The update at each step is
\begin{align}
    \alpha^{(i+1)} &= \alpha^{(i)} - \eta \nabla_\alpha L_{I^{(V)}_t}\left(\theta_{t} - \gamma \sigma(\alpha^{(i)}) \nabla L_{I_t^{(T)}}(\theta_{t})\right)
    \nonumber \\
    \beta^{(i+1)} &= \sigma(\alpha^{(i+1)}) \label{eqn:pola-gd}
\end{align}
where $\sigma(\alpha) = \frac{1}{1+e^{-\alpha}}$ is the sigmoid function.

\begin{algorithm}[h]
\caption{POLA\label{alg:pola}}

\hspace*{\algorithmicindent} \textbf{Input:} Historical sequence length $n$, forecast horizon $m$, online batch size $b$, maximum learning rate $\gamma$, moving average window q \\
\hspace*{\algorithmicindent}
\textbf{Additional input for POLA-FS:} Finite set of candidate learning rates $\mathcal{C}$ \\
\hspace*{\algorithmicindent}
\textbf{Additional input for POLA-GD:} Number of update steps $k$, learning rate $\eta$ \\
\hspace*{\algorithmicindent} \textbf{Output:} Predictions $\hat{y}_t$ for $t=1,\dots,T$
\begin{algorithmic}[1]

\Procedure{POLA}{}
\State Initialize $S=0$, $B=[]$
\For{$t=1:T$}
    \State Predict $\hat{y}_t = f(x_t, \theta_t)$
    \If{$S==b$} \Comment{batch update}
        \State Collect indices $I_t = \{t-n-b+1, \dots, t-n\}$ of update batch 
        \State Create meta-training set $U_t^{(T)}$ and meta-validation set $U_t^{T}$ as in Equation~\ref{eqn:split}
        \State Optimize for $\beta^{(V)}_{t+1}$ as in Equation~\ref{eqn:pola-fs} for POLA-FS and Equation~\ref{eqn:pola-gd} for POLA-GD
        \State B.append$\left(\beta_{t+1}^{(V)}\right)$
        \State Set $\beta_{t+1} = \frac{1}{q}\sum_{i=1}^{q} B[-i]$ 
        \State Update $\theta_{t+1} = \theta_{t} - \gamma \beta_{t+1} \nabla L_{I_t}(\theta_{t})$
        \If{$\beta_{t+1} > 0$}
        \State $S=0$ \Comment{reset counter}
        \EndIf
    \Else
        \State $\theta_{t+1} = \theta_t$
        \State $S = S+1$ \Comment{update counter}
    \EndIf
\EndFor
\EndProcedure

\end{algorithmic}
\end{algorithm}

\section{EXPERIMENTS}

We test our proposed POLA on 2 publicly-available time series datasets with 3 commonly-used recurrent network structures: Recurrent Neural Network (RNN), Long Short-Term Memory network (LSTM), and Gated Recurrent Unit network (GRU). We use recurrent networks with 10 neurons. For all datasets, we allow a pre-training period with 700 samples for hyperparameter tuning and use prequential or interleaved-test-then-train evaluation scheme on the rest of the samples, denoted as the online phase, as is common in streaming data applications. We use an online batch size $b=10$ and evaluate for multi-step prediction, which is more difficult than one-step prediction since the model needs to learn temporal patterns instead of simply aggregating the most recently observed values~\cite{taieb2011nn5}. We evaluate prediction performance with normalized root-mean-square error (RMSE) where RMSE is scaled by the standard deviation of the data to allow comparison across datasets. All results presented are averaged across runs over 10 different seeds.

We compare with a variety of time series and neural network methods. We include a popular exponential smoothing method Holt-Winters which is shown to be a strong baseline for time series forecasting~\cite{hewalage2020rnn}, and an extreme-learning machine method for time series OR-ELM~\cite{park2017orelm}. For neural network methods, the Pre-trained RNN is trained with offline configurations of 500 epochs with learning rate 0.1 and batch size 32 on the pre-training samples and not updated in the online phase. Follow The Leader (FTL) RNN is retrained every $b$ steps with all historical data. We limit the number of batches to match that in the pre-training phase to constrain online training time. MAML is a meta-learning method that aims to learn parameters offline which can generalize better to unseen data distributions, and we apply the variation of MAML for sequential data in \cite{Nagabandi2019LearningTA} to RNN. We also include 3 gradient-descent online methods applied on recurrent networks. The first uses SGD with constant learning rate, the second uses RMSprop~\cite{rmsprop} which is a popular algorithm for training neural networks with element-wise adaptive learning rates, and the third is a weighted gradient (WG) algorithm that adapts the learning rate of SGD based on whether the current sample is an outlier or change point~\cite{guo2016robust}. Since WG is designed for one-step prediction~\cite{guo2016robust}, we implement WG to adapt the learning rate using only the first-step prediction. The maximum learning rate $\gamma$ for these gradient-descent methods including POLA is determined by hyperparameter tuning on the set $\mathcal{C}=\{1, 0.1, 0.01, 0.001, 0.0001, 0\}$ by splitting the pre-training samples into two-thirds for training and one-thirds for validation. This is the same set $\mathcal{C}$ used in POLA-FS. Hyperparameters for POLA-GD is set to $k=3$ and $\eta=0.1$, and the POLA moving average window $q$ is tuned on $\{1,3,5,7,9\}$ on the pre-training samples.

\subsection{Dataset Description}

\textbf{Sunspot:} The monthly sunspot number dataset\footnote{\url{http://www.sidc.be/silso/datafiles}} contains the arithmetic mean of the daily sunspot number over all days of each calendar month from January, 1749 to July, 2020. Each cycle in the sunspot number series is approximately 11 years. We set the length of historical data $n=48$ which corresponds to 4 years as used by \cite{miranian2013fuzzy} for annual data, and forecast length $m=5$ which is 5 months.

\textbf{Household Power Consumption:} The household power consumption dataset\footnote{\url{https://archive.ics.uci.edu/ml/datasets/Individual+household+electric+power+consumption}} is a multivariate dataset that records the electricity consumption of a household from December 16, 2006 to November 26, 2010. Due to missing values, we take daily averages and forward padded any remaining missing values. The measurements for global active power, global intensity and voltage are used for prediction. Other variables are not included due to sparse or low readings. We set the length of historical data $n=28$ which is 4 weeks, and forecast length $m=3$ which is 3 days. 

These datasets are known to exhibit concept drift~\cite{miranian2013fuzzy, Zhang2019ABACUSUM}.

\subsection{Online Prediction Performance}

Table~\ref{tab:rnn} presents the prediction performance of RNN and other time series models. Holt-Winters and OR-ELM generally perform worse than RNN methods, likely because RNN has better modeling capacity for complex temporal patterns. MAML has the highest RMSE, possibly because the small amount of pre-training data does not have sufficiently diverse examples of data distribution changes to generalize to those in the online phase. Both Pre-trained and FTL RNN incur higher errors than the gradient-descent online RNNs, which indicate the presence of concept drift between the pre-training and online phase, and also within the online phase. Both versions of POLA have the best performances for both datasets, showing that the automatic learning rate scheduler is effective in helping the model adapt. 

\begin{figure}
\centering
\begin{subfigure}[b]{.7\linewidth}
    \centering
    \includegraphics[width=\linewidth]{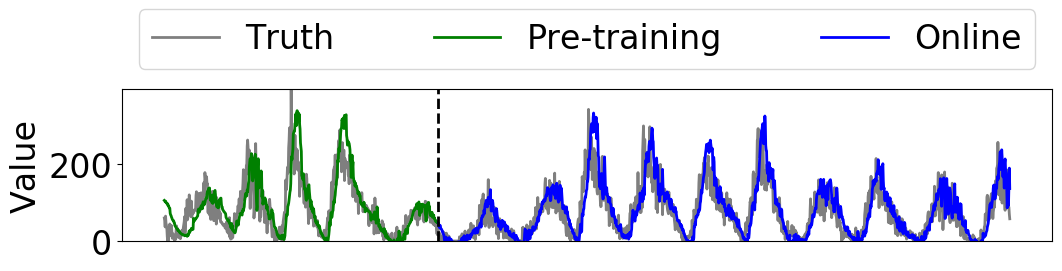}
\end{subfigure}

\begin{subfigure}[b]{.7\linewidth}
    \centering
    \includegraphics[width=\linewidth]{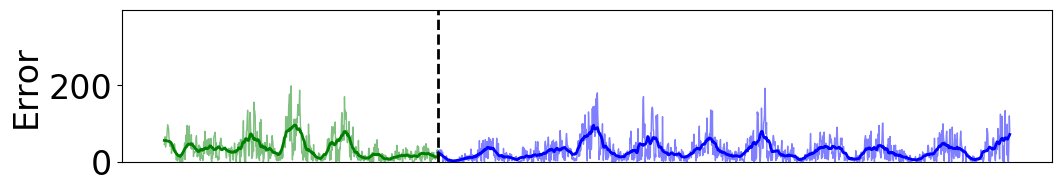}
\end{subfigure}

\begin{subfigure}[b]{.7\linewidth}
    \centering
    \includegraphics[width=\linewidth]{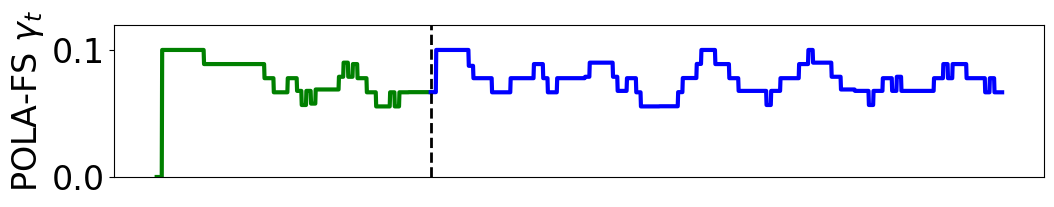}
\end{subfigure}

\begin{subfigure}[b]{.7\linewidth}
    \centering
    \includegraphics[width=\linewidth]{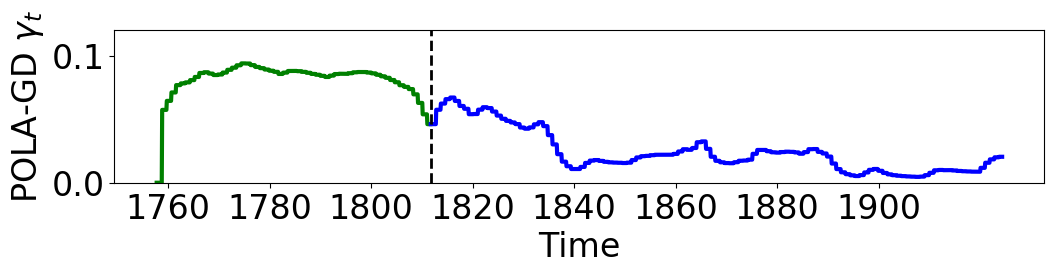}
\end{subfigure}
\caption[short]{Predictions (5-step ahead), errors (smoothed across 2 years) and POLA learning rates for Sunspot dataset.\label{fig:sunspot}}
\vspace{-2em}
\end{figure}

The additional meta-learning procedure is expected to increase computation time. Approximately in seconds on the Sunspot dataset with 3259 observations, Online-SGD and Online-RMSprop takes 3.5s, WG and POLA-FS takes 6.5s, POLA-GD takes 40s with k=1 and 250s with k=3 update steps. Figure~\ref{fig:sunspot} plots the POLA-FS predictions, errors and learning rates. POLA-GD predictions and errors are visually similar. We observe different behaviors in the learning rate $\gamma_t$. From Equation~\ref{eqn:pola-gd} for POLA-GD, the rate of change of $\gamma_t$ depends directly on $\eta$, hence we see slower-moving changes with a low $\eta$. The sigmoid shape also means that $\sigma(\alpha)$ will tend to be near 0 or 1, which is reflected through POLA-GD $\gamma_t$ being near 0 or 0.1.

We evaluate online recurrent network methods on other architectures in Table~\ref{tab:otherrnn}. POLA achieve comparable or better performance than competing methods. The improved performance of Online-RMSprop in LSTM compared to the other architectures could be due to the larger number of model parameters, such that element-wise learning rates can provide localized adaptation. However while POLA is consistently better or on par with baseline Online-SGD, RMSprop is inconsistent and in some cases performs much worse. 

\begin{table}[h]
\centering

{\setlength\doublerulesep{0.4pt}   
\begin{tabular}{lll}
\toprule[1pt]\midrule[0.3pt]
\textbf{METHOD}      & \multicolumn{2}{c}{\textbf{NORMALIZED RMSE}}                                \\ \cmidrule{2-3}
                     & Sunspot           & Power  \\ \midrule
Holt-Winters         & 0.991             & NA   \\
OR-ELM               & 0.822             & NA \\ \midrule
Pre-trained          & 0.572             & 0.816 \\
FTL$^*$              & 0.572             & 0.820\\
MAML                 & 1.295             & 1.023 \\
Online-SGD           & 0.552             & 0.775 \\
Online-RMSprop       & 0.536             & 0.809 \\
WG                   & 0.552             & NA \\ \midrule
POLA-FS              & \underline{0.532} \rpm 0.002 & \textbf{0.769} \rpm 0.003 \\
POLA-GD              & \textbf{0.500} \rpm 0.002   & \underline{0.773} \rpm 0.005 \\
\midrule[0.3pt]\bottomrule[1pt]
\end{tabular}
}  
\caption{Prediction performance of RNN and time series models. Results of univariate methods on multivariate dataset are marked NA. $^*$ FTL retrains with entire historical data. \label{tab:rnn}}
\vspace{-2em}
\end{table}

\begin{table}[h]
\centering

{\setlength\doublerulesep{0.4pt}   
\begin{tabular}{llll}
\toprule[1pt]\midrule[0.3pt]
\textbf{MODEL}   & \textbf{METHOD}  & \multicolumn{2}{c}{\textbf{NORMALIZED RMSE}}  \\ \cmidrule{3-4}
                    &                  & Sunspot           & Power  \\ \midrule
LSTM                & Online-SGD     & 0.532             & 0.821 \\
                    & Online-RMSprop & \underline{0.517} & \textbf{0.794} \\
                    & WG               & 0.532             & NA \\ \cmidrule{2-4}
                    & POLA-FS        & 0.534 \rpm 0.009         & \underline{0.802} \rpm 0.005 \\
                    & POLA-GD        & \textbf{0.512} \rpm 0.006   & 0.806 \rpm 0.071 \\ 
                    \midrule
GRU                 & Online-SGD     & 0.526             & \textbf{0.768} \\
                    & Online-RMSprop & 0.521             & 0.786 \\
                    & WG               & 0.526             & NA \\ \cmidrule{2-4}
                    & POLA-FS        & \underline{0.508} \rpm 0.002 & \underline{0.769} \rpm 0.003 \\
                    & POLA-GD        & \textbf{0.489} \rpm 0.002   & \textbf{0.768} \rpm 0.003 \\ 
\midrule[0.3pt]\bottomrule[1pt]                    
\end{tabular}
}  
\caption{Prediction performance of recurrent network models. Results of univariate methods on multivariate dataset are marked NA.\label{tab:otherrnn}}
\vspace{-1em}
\end{table}

\begin{table}[h]
\centering

{\setlength\doublerulesep{0.4pt}   
\begin{tabular}{llll}
\toprule[1pt]\midrule[0.3pt]
\textbf{\# STEPS}   & \textbf{LEARNING}  & \multicolumn{2}{c}{\textbf{NORMALIZED RMSE}}  \\ \cmidrule{3-4}
(k)                 & \textbf{RATE $(\eta)$}           & Sunspot           & Power  \\ \midrule
1 & 0.1 & 0.515 & 0.773 \\
2 & 0.1 & 0.504 & 0.772 \\
3 & 0.1 & 0.500 & 0.773 \\
1 & 0.01 & 0.525 & 0.777 \\
2 & 0.01 & 0.520 & 0.776 \\
3 & 0.01 & 0.516 & 0.775 \\
\midrule[0.3pt]\bottomrule[1pt]                    
\end{tabular}
}  
\caption{Prediction performance of POLA-GD RNN learned with different hyperparameters.\label{tab:pola-gd}}
\vspace{-1em}
\end{table}

\begin{figure}[h!]
    \centering
    \includegraphics[width=\linewidth]{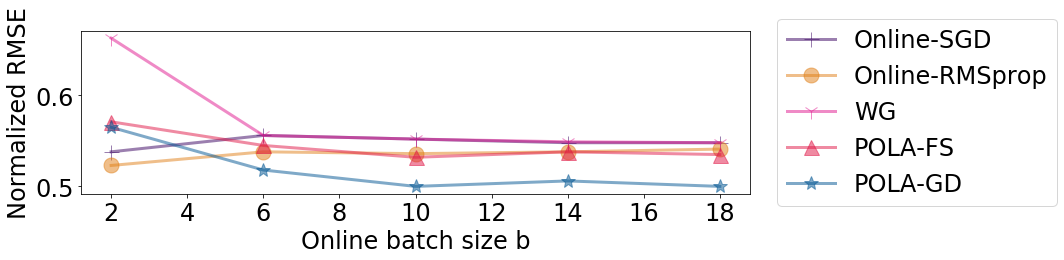}
    \caption{Normalized RMSE across different online batch sizes $b$ with RNN model on Sunspot dataset.}
    \label{fig:sensitivity_b}
    \vspace{-2em}
\end{figure}

\subsection{Sensitivity Analysis}

We conduct sensitivity analysis on the hyperparameters of POLA-GD and the online batch size $b$. From Table~\ref{tab:pola-gd}, we see that the prediction performance of POLA-GD RNN can depend on the choice of hyperparameters $k$ and $\eta$. Overall, a larger number of update steps $k$ tend to give better performance as the optimization can search for a better solution of $\beta_{t+1}^{(V)}$.

Figure~\ref{fig:sensitivity_b} shows the performance of online RNN methods using different online batch size $b$ for $b\in \{2, 6, 10, 14, 18\}$. POLA has comparable or better performance than competing methods at $b\geq 10$. Since each batch of samples is split into two at the meta-stage, a larger $b$ helps to estimate the learning rate more accurately.

\section{CONCLUSION}

In this paper, we propose POLA to automatically adapt the SGD learning rate to adjust recurrent neural network models for predicting real-time in dynamic environments. The learning rate is estimated using a novel meta-learning procedure that assimilates the interleaved-test-then-train evaluation. POLA demonstrates overall comparable or better predictive performance over other competing methods across multiple datasets and network architectures.

\bibliographystyle{IEEEbib}
\bibliography{references}

\end{document}